# Generating lyrics with variational autoencoder and multi-modal artist embeddings

**Olga Vechtomova, Hareesh Bahuleyan, Amirpasha Ghabussi, Vineet John**
University of Waterloo, ON, Canada
{ovechtom,hpallika,aghabuss,vineet.john}@uwaterloo.ca

## Abstract

We present a system for generating song lyrics lines conditioned on the style of a specified artist. The system uses a variational autoencoder with artist embeddings. We propose the pre-training of artist embeddings with the representations learned by a CNN classifier, which is trained to predict artists based on MEL spectrograms of their song clips. This work is the first step towards combining audio and text modalities of songs for generating lyrics conditioned on the artist's style. Our preliminary results suggest that there is a benefit in initializing artists' embeddings with the representations learned by a spectrogram classifier.

## 1 Introduction

Outputs of neural generative models can serve as an inspiration for artists, writers and musicians when they create original artwork or compositions. In this work we explore how generative models can assist songwriters and musicians in writing song lyrics. In contrast to systems that generate lyrics for an entire song, we propose to generate suggestions for lyrics lines in the style of a specified artist. The hope is that unusual and creative arrangements of words in the generated lines will inspire the songwriter to create original lyrics. Conditioning the generation on the style of a specific artist is done in order to maintain stylistic consistency of the suggestions. Such use of generative models is intended to augment the natural creative process when an artist may be inspired to write a song based on something they have read or heard.

We use the variational autoencoder (VAE) [1] with Long Short Term Memory networks (LSTMs) as encoder and decoder, and a trainable artist embedding, which is concatenated with the input to every time step of the decoder LSTM. The advantage of using the VAE for creative tasks, such as lyrics generation, is that once the VAE is trained, any number of lines can be generated by sampling from the latent space. The unique style of each musician is a combination of their musical style and the style expressed through their lyrics. We therefore compare randomly initialized trainable artist embeddings with embeddings pre-trained by a convolutional neural network (CNN) classifier optimized to predict the artist based on MEL spectrograms of 10-second song clips.

There are a large number of approaches towards poetry generation. Some approaches focus on such characteristics as rhyme and poetic meter [2], while others on generating poetry in the style of a specific poet [3]. In [4] the authors propose image-inspired poetry generation. The approach of using style embeddings in controlled text generation is not new, and has been explored in generating text conditioned on sentiment [5, 6] and persona-conditioned responses in dialogue systems [7]. To our knowledge there has been no prior work on using music audio and text modalities to generate song lyrics. We explore whether artist embeddings learned based on music audio are useful in generating lyrics in the style of a given artist. Our preliminary results suggest that there is some benefit in using multi-modal embeddings for conditioned lyrics generation.

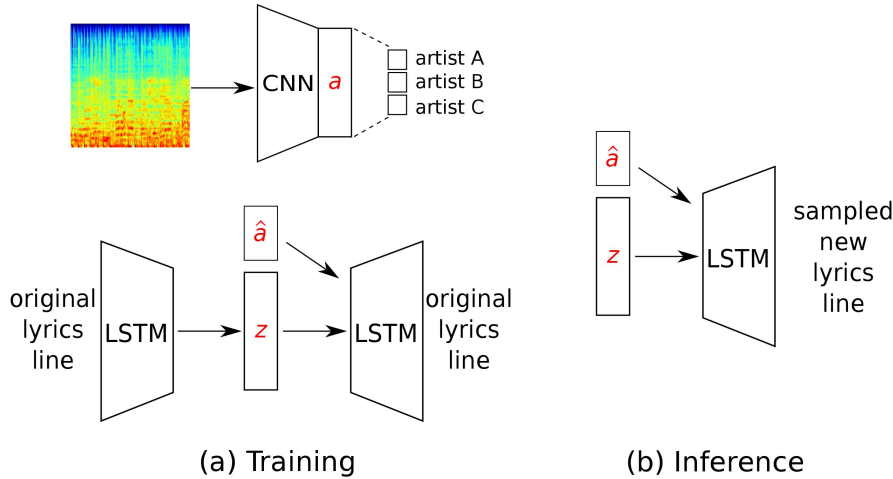

(a) Training    (b) Inference

Figure 1: Overview of our approach - First, a CNN is implemented to classify artists based on spectrogram images, thereby learning artist embeddings. Then, a VAE is trained to reconstruct lines from song lyrics, conditioned on the pre-trained artist embeddings. At inference time, in order to generate lyrics in the style of a desired artist, we sample $z$ from the latent space and decode it conditioned on the embedding of that artist.

## 2 Model and Experiments

We collected a dataset of lyrics by seven artists, one from each of the following genres: Art Rock, Electronic, Industrial, Classic Rock, Alternative, Hard Rock, and Psychedelic Rock. Each of the selected artists has a distinct musical and lyrical style, and a large catalogue of songs, spanning many years. In total our dataset contains 34,000 lines of song lyrics.

To obtain pre-trained artist embeddings, we split the waveform audio of the artists' songs into 10-second clips, and transformed them into MEL spectrograms[1]. The dataset consists of 21,235 spectrograms. Next, a VGG16 [8] pre-trained CNN classifier was trained to predict artists based on spectrograms[2]. The classifier achieved an accuracy of 83% on the test set. The last hidden layer of the classifier was used to initialize the artist embeddings of the lyrics VAE. The VAE is trained to perform the task of sentence reconstruction on the lyrics dataset. At inference time, we sample data points from the learned latent space, and pass them to the decoder together with the embedding of the artist, in whose style we want to generate lyrics. Two variants of this model were evaluated: VAE+audioT and VAE+audioNT, with trainable and non-trainable artist embeddings, respectively.

For the baseline we implemented the VAE model with randomly initialized artist embeddings: VAE+randT (trainable) and VAE+randNT (non-trainable), and VAE with artist embeddings as one-hot encodings (VAE+onehot). All VAE models were trained by annealing the coefficient of the KL cost up to 3000 iterations and a decoder input word dropout of 0.5 [9, 10]. The encoder is a bi-directional LSTM with 100 hidden units. The dimension of the artist embedding vector was set to 50 in all VAEs except for VAE+onehot. We used 300-dimension word2vec embeddings pre-trained on a large corpus of song lyrics (2.5M lines).

The spectrogram CNN model consists of a CNN base model, which is the VGG-16 and uses pre-trained ImageNet weights [8]. The CNN base is followed by three fully-connected layers (512,128,50 units) with 30% dropout. The model was trained for 20 epochs. Classification accuracy on the test set (80/10/10 training/validation/test split) was 83%. While creating the data splits, we ensured that the audio clips for the same song remained in the same set.

Examples of the generated lyrics are given in Table 1. Poems composed by the first author from the lines generated by VAE+audioNT have been accepted as artwork at the NeurIPS 2018 Workshop for Creativity and Design[3].

---

[1] https://www.kaggle.com/vinvinvin/high-resolution-mel-spectrograms/notebook
[2] https://github.com/pashapanther/deep-music-genre-classification
[3] http://www.aiartonline.com/community/olga-vechtomova/

| Electronic | Art Rock |
|---|---|
| like shackles of the eternal night | love can drown your heart |
| oh i want to shake the sun | no way to heaven where she stands |
| black obsession is wearing in your soul | when the shadows were young |
| Industrial | Alternative |
| every inch of reptile in your head | i'm drifting away from the sea |
| just when the jagged sound | for superior betrayal |
| i watch me get into my skin | forevermore he held the earth |

Table 1: Lyrics lines generated by VAE+audioNT

## 3 Evaluation and Results

To quantitatively evaluate whether the generated lyrics adhere to the style of the artist they were conditioned on, we trained a CNN classifier [11][4] on the original lyrics of the selected seven artists. This is a commonly used approach to evaluate the style attribute of generated texts, e.g. in style transfer [12]. The results are presented in Table 2. The accuracy on the original lyrics is 60%, and the majority baseline is 17.7%. The results are presented in Table 2. The VAE+audioNT model received the highest style classification accuracy of 42%, which suggests that there is some benefit in pre-training artist embeddings on spectrogram images.

The performances of VAE+randT and VAE+randNT are somewhat variable between training instances due to the random initialization of the artist embeddings. This is evident from the fact that after each training instance the embedding vectors of different pairs of artists have the highest cosine similarity. To account for this variability, we trained five instances of each model and averaged their evaluation results. The VAE+randT and VAE+randNT results presented in Tables 2 and 3 are averaged over five training instances.

| Model | Accuracy |
|---|---|
| VAE+onehot | 0.266 |
| VAE+randT | 0.368 |
| VAE+randNT | 0.396 |
| VAE+audioT | 0.361 |
| VAE+audioNT | **0.420** |

Table 2: Style classification accuracy on the generated lyric lines

We also trained a Kneser-Ney smoothed trigram language model [13] on the corpus of each artist's lyrics, and then used each of the seven artists' language models to score the lyrics generated for any given artist. The intuition is that if the model successfully generates lyrics in the style of a given artist, then that artist's language model should result in the lowest negative log-likelihood value. In VAE+audioNT, for six out of seven artists, the lowest negative log likelihood values were given by the model trained on the same artist's lyrics, which suggests that our model generates lyrics in the style of a specified artist. Table 3 contains the results of the VAE model with non-trainable randomly initialized artist embeddings (VAE+randNT), while Table 4 contains the results of VAE+audioNT.

While the above metrics evaluate how well the models generate lines in the style of an artist, the perfect scores would be obtained by systems that simply learn to reproduce the original lines, whereas what we want are new lines that are "inspired" by the artist's lyrics, but are not verbatim copies. The number of verbatim copies among all evaluated models was very low (2-3%). Also, all models generated diverse lines: 98%-99% of lines are unique.

A small-scale human evaluation was conducted to assess how close the generated lines are to the style of a specific artist. We recruited three annotators, one of whom was familiar with the selected artists in Electronic and Classic Rock genres, and two annotators were familiar with one artist each. We obtained 100 samples of lyrics lines generated by each of the four VAE models conditioned on each artist, shuffled and presented them to each evaluator. The evaluators were asked to select the lines that resemble the style of the given artist. The results (Table 5) indicate that except for one case, VAE+audioT and VAE+audioNT generated the most lines in the style of the given artist, although the differences are rather small. Cohen's kappa between the pairs of annotators was low, which can be explained by the subjective nature of judging an artist's style.

---
[4] https://github.com/dennybritz/cnn-text-classification-tf

| Artist genre | Language model | | | | | | |
|---|---|---|---|---|---|---|---|
| | AR | E | I | CR | A | HR | PR |
| Art Rock (AR) | **16.9** | 17.44 | 17.32 | 17.55 | 17.79 | 17.89 | 17.5 |
| Electronic (E) | 17.49 | **16.23** | 16.63 | 17.34 | 17.48 | 17.47 | 17.34 |
| Industrial (I) | 17.37 | 16.85 | **15.68** | 17.42 | 17.3 | 17.51 | 17.32 |
| Classic Rock (CR) | 17.66 | 17.39 | 17.24 | **16.99** | 17.8 | 17.89 | 17.48 |
| Alternative (A) | 17.47 | 17.18 | 16.82 | 17.43 | **16.82** | 17.54 | 17.23 |
| Hard Rock (HR) | 16.83 | 16.54 | 16.6 | 16.82 | 16.91 | **16.22** | 16.86 |
| Psychedelic Rock (PR) | 17.1 | 17.14 | 17.12 | 17.19 | 17.43 | 17.53 | **16.29** |

Table 3: Negative log-likelihood values for the lyrics generated by **VAE+randNT**. The language models were trained on the original lyrics of artists.

| Artist genre | Language model | | | | | | |
|---|---|---|---|---|---|---|---|
| | AR | E | I | CR | A | HR | PR |
| Art Rock (AR) | **15.5** | 15.95 | 16.19 | 16.04 | 16.29 | 16.43 | 15.81 |
| Electronic (E) | 16.38 | **15.08** | 15.89 | 16.36 | 16.38 | 16.31 | 16.36 |
| Industrial (I) | 16.47 | 16.01 | **15.16** | 16.66 | 16.47 | 16.61 | 16.37 |
| Classic Rock (CR) | 17.09 | 16.86 | 16.78 | **16.32** | 17.07 | 17.07 | 16.88 |
| Alternative (A) | 17.74 | 17.3 | 16.92 | 17.77 | 16.95 | 17.67 | 17.35 |
| Hard Rock (HR) | 17.49 | 17.04 | 17.07 | 17.13 | 17.63 | **16.7** | 17.28 |
| Psychedelic Rock (PR) | 17.07 | 17.23 | 17.15 | 17.27 | 17.22 | 17.24 | **16.37** |

Table 4: Negative log-likelihood values for the lyrics generated by **VAE+audioNT**. The language models were trained on the original lyrics of artists (smaller values are better).

| Model | Electronic | | Classic Rock | |
|---|---|---|---|---|
| | Annotator A | Annotator B | Annotator A | Annotator C |
| VAE+onehot | 0.79 | 0.29 | 0.67 | 0.34 |
| VAE+randT | **0.8** | 0.35 | 0.67 | 0.3 |
| VAE+audioT | 0.79 | 0.33 | **0.7** | 0.32 |
| VAE+audioNT | 0.73 | **0.37** | 0.6 | **0.4** |

Table 5: Manual evaluation results (ratio of selected lines out of 100 generated lines per artist).

## 4 Conclusions and Future Work

Our initial results are promising and suggest that pre-training artist embeddings on music spectrograms helps to condition lyric generation on the artist's style. Since artist embeddings are pre-trained using a separate model, their meaning is not known to the VAE. However, the difference between artist embeddings is meaningful, as it reflects the difference between their musical styles. Our approach is based on the assumption that artists with similar musical styles, and hence, similar audio-derived embeddings, have more similar lyrical styles than artists that are very different musically.

In future work, we plan to evaluate other models for pre-training of artist embeddings, for example spectrogram autoencoders. We will also explore other approaches to learn multi-modal representations, e.g. [14] and adversarial approaches.

## References

[1] Diederik P Kingma and Max Welling. Auto-encoding variational Bayes. In *Proceedings of the International Conference on Learning Representations*, 2014.

[2] Xingxing Zhang and Mirella Lapata. Chinese poetry generation with recurrent neural networks. In *Proceedings of the 2014 Conference on Empirical Methods in Natural Language Processing (EMNLP)*, pages 670–680, 2014.

[3] Aleksey Tikhonov and Ivan P Yamshchikov. Guess who? multilingual approach for the automated generation of author-stylized poetry. *arXiv preprint arXiv:1807.07147*, 2018.

[4] Wen-Feng Cheng, Chao-Chung Wu, Ruihua Song, Jianlong Fu, Xing Xie, and Jian-Yun Nie. Image inspired poetry generation in xiaoice. *arXiv preprint arXiv:1808.03090*, 2018.

[5] Zhiting Hu, Zichao Yang, Xiaodan Liang, Ruslan Salakhutdinov, and Eric P. Xing. Toward controlled generation of text. In *Proceedings of the 34th International Conference on Machine Learning*, pages 1587–1596, 2017.

[6] Zhenxin Fu, Xiaoye Tan, Nanyun Peng, Dongyan Zhao, and Rui Yan. Style transfer in text: Exploration and evaluation. In *AAAI*, pages 663–670, 2018.

[7] Jiwei Li, Michel Galley, Chris Brockett, Georgios Spithourakis, Jianfeng Gao, and Bill Dolan. A persona-based neural conversation model. In *Proceedings of the 54th Annual Meeting of the Association for Computational Linguistics (Volume 1: Long Papers)*, pages 994–1003. Association for Computational Linguistics, 2016.

[8] Karen Simonyan and Andrew Zisserman. Very deep convolutional networks for large-scale image recognition. *arXiv preprint arXiv:1409.1556*, 2014.

[9] Samuel R. Bowman, Luke Vilnis, Oriol Vinyals, Andrew Dai, Rafal Jozefowicz, and Samy Bengio. Generating sentences from a continuous space. In *Proceedings of the 20th SIGNLL Conference on Computational Natural Language Learning*, pages 10–21, 2016.

[10] Hareesh Bahuleyan, Lili Mou, Olga Vechtomova, and Pascal Poupart. Variational attention for sequence-to-sequence models. *Proceedings of the 27th International Conference on Computational Linguistics (COLING)*, 2018.

[11] Yoon Kim. Convolutional neural networks for sentence classification. *arXiv preprint arXiv:1408.5882*, 2014.

[12] Tianxiao Shen, Tao Lei, Regina Barzilay, and Tommi Jaakkola. Style transfer from non-parallel text by cross-alignment. In *NIPS*, pages 6833–6844, 2017.

[13] Reinhard Kneser and Hermann Ney. Improved backing-off for m-gram language modeling. In *ICASSP*, 1995.

[14] Hongru Liang, Haozheng Wang, Jun Wang, Shaodi You, Zhe Sun, Jin-Mao Wei, and Zhenglu Yang. Jtav: Jointly learning social media content representation by fusing textual, acoustic, and visual features. *arXiv preprint arXiv:1806.01483*, 2018.